\pdfoutput=1

\documentclass[11pt]{article}

\usepackage[]{EMNLP2023}

\usepackage{times}
\usepackage{latexsym}

\usepackage[T1]{fontenc}
\usepackage[utf8]{inputenc}

\usepackage[russian,english]{babel}

\usepackage{microtype}

\usepackage{times}
\usepackage{amssymb}
\usepackage{latexsym}
\usepackage{booktabs}
\usepackage{xspace}
\usepackage{xcolor}
\usepackage{enumitem}

\usepackage{graphicx}

\usepackage[disable]{todonotes}

\usepackage{hyperref}
\usepackage{multirow}
\usepackage{tabularx}
\usepackage{subcaption}
\usepackage{newtxmath}

\newcolumntype{L}{>{\raggedright\arraybackslash}X}
\newcolumntype{R}{>{\raggedleft\arraybackslash}X}
\newcolumntype{C}{>{\raggedcenter\arraybackslash}X}

\definecolor{royalpurple}{RGB}{136,18,255}
\definecolor{royalblue}{RGB}{0,102,204}

\newcommand{\matt}[2][]{\todo[inline,size=\scriptsize,color=teal!40,#1]{#2}}

\newcommand{\ersatz}{\textsc{Ersatz}\xspace}

\newcommand{\auxiliary}{\textsc{Auxiliary}\xspace}
\newcommand{\gender}{\textsc{Gender}\xspace}
\newcommand{\formality}{\textsc{Formality}\xspace}
\newcommand{\inflection}{\textsc{Inflection}\xspace}
\newcommand{\animacy}{\textsc{Animacy}\xspace}

\newcommand{\ctxpro}{\textsc{ctxpro}\xspace}

\renewcommand{\check}{\ensuremath{\checkmark}}

\usepackage{inconsolata}

\title{Identifying Context-Dependent Translations for Evaluation Set Production}

\author{Rachel Wicks$^{1, 2}$ \and Matt Post$^{1-3}$\\
  ${}^1$Human Language Technology Center of Excellence, Johns Hopkins University\\
  ${}^2$Center of Language and Speech Processing, Johns Hopkins University \\
  ${}^3$Microsoft \\
  \texttt{rewicks@jhu.edu, mattpost@microsoft.com}}

\begin{document}
\maketitle
\begin{abstract}

A major impediment to the transition to context-aware machine translation is the absence of good evaluation metrics and test sets.
Sentences that require context to be translated correctly are rare in test sets, reducing the utility of standard corpus-level metrics such as COMET or BLEU.
On the other hand, datasets that annotate such sentences are also rare, small in scale, and available for only a few languages.
To address this, we modernize, generalize, and extend previous annotation pipelines to produce \ctxpro, a tool that identifies subsets of parallel documents containing sentences that require context to correctly translate five phenomena: gender, formality, and animacy for pronouns, verb phrase ellipsis, and ambiguous noun inflections.
The input to the pipeline is a set of hand-crafted, per-language, linguistically-informed rules that select contextual sentence pairs using coreference, part-of-speech, and morphological features provided by state-of-the-art tools.
We apply this pipeline to seven languages pairs (EN into and out-of DE, ES, FR, IT, PL, PT, and RU) and two datasets (OpenSubtitles and WMT test sets), and validate its performance using both overlap with previous work and its ability to discriminate a contextual MT system from a sentence-based one.
We release the \ctxpro pipeline and data as open source.\footnote{\url{https://github.com/rewicks/ctxpro}}
\todo{mention the ctxpro name here}
\end{abstract}

\section{Introduction}
\label{sec:introduction}

\begin{table}[t]
    \centering
\setlength{\tabcolsep}{2pt}
\begin{tabular}{@{}lcccccl@{}}
\toprule
 & 
\multicolumn{1}{l}{\rotatebox{90}{\small\gender}} & 
\multicolumn{1}{l}{\rotatebox{90}{\small\auxiliary}} & 
\multicolumn{1}{l}{\rotatebox{90}{\small\inflection}} &
\multicolumn{1}{l}{\rotatebox{90}{\small\formality}} &
\multicolumn{1}{l}{\rotatebox{90}{\small\animacy}} &
\textsc{langs} \\
\midrule
M\"uller et al. & \check &  &  &  & & de \\
Lopes et al. & \check &  &  &  & & fr \\
Voita et al. & \check & \check & \check & \check & & ru \\
Nadejde et al. &  &  &  & * & & de, es, fr, hi, it, ja \\
\multirow{2}{*}{Currey et al.} & \multirow{2}{*}{$\dagger$} & & & & & ar, fr, de, hi, it, pt,\\
    & & & & & & ru, es \\

\midrule
\textbf{This work} & \check & \check & \check & \check & \check & de, fr, ru, pl, pt, it, es \\
\bottomrule
\end{tabular}
    \caption{This work expands evaluation set coverage to new document phenomena and languages. (*) Note that \citet{nadejde-etal-2022-cocoa} does not include contextual information. ($\dagger$) \citet{currey-etal-2022-mt} focuses on natural, rather than grammatical, gender.}
    \label{tab:thiswork}
    \vspace{-6mm}
\end{table}

\begin{table*}[ht]
    \centering

\begin{tabular}{lp{6.5cm}p{6.5cm}}

\toprule
 & \multicolumn{1}{c}{English} & \multicolumn{1}{c}{Target} \\
 \midrule
 
\multicolumn{1}{l}{\multirow{2}{*}{\auxiliary}} & I just figured you need to know. \textit{And now you \textbf{do.}} & 
\texttt{(fr)} Je pensais que tu méritais de savoir. \textit{Et maintenant tu \textbf{sais.}} \\

\multicolumn{1}{l}{} & I can't lose my voice. \textit{You \textbf{won't. }}& 
\texttt{(pl)} Nie mogę stracić głosu. \textit{Nie \textbf{stracisz.}} \\
\midrule

\multicolumn{1}{l}{\inflection} & Mostly work with the Knicks right now. 
\textit{And other \textbf{athletes.}} &

\texttt{(ru)} \foreignlanguage{russian}{В основном работаю с ``Никс''. \textit{И с другими \textbf{спортсменами.}}}\\
\midrule

\multicolumn{1}{l}{\multirow{2}{*}{\gender}} & You think migraines are a sign of weakness, don't want anyone to know. \textit{I used to get \textbf{them}, too. }& 
\texttt{(it)} Lei pensa che le emicranie siano segno di debolezza, e non vuole che si sappia. \textit{\textbf{Le} prendevo anch'io.} \\

\multicolumn{1}{l}{} & This pain? \textit{I long for \textbf{it.}} & 
\texttt{(pt)} A dor? \textit{Anseio por \textbf{ela.}} \\
\midrule

\multicolumn{1}{l}{\multirow{2}{*}{\animacy}} & Et il y a eu cette rose aussi pour toi. \textit{Tu sais, \textbf{elle} se distingue des autres.} &
\texttt{(en)} Also, uh, this rose came for you. \textit{You know, \textbf{it} stands out in front of all the others.} \\

\multicolumn{1}{l}{} & La felicidad es un mito. \textit{Y vale la pena luchar por \textbf{ella.}} &
\texttt{(en)} Happiness is a myth. \textit{And \textbf{it's} worth fighting for.} \\
\midrule

\multirow{2}{*}{\formality} & \textit{We'll call \textbf{you} if something happens, huh?} & 

\texttt{(de)} \textit{Wir rufen \textbf{euch} an, wenn etwas passiert.} \\
 & \textit{Well, uh, I was an obstetrician before, and I most definitely owe \textbf{you.}} & 
 
 \texttt{(es)} \textit{Bueno, era obstetra antes, y definitivamente se \textbf{los} debo.} \\
 \bottomrule
\end{tabular}

    \caption{An example of the extracted ambiguities with their preceding contexts for each language pair. The ambiguous sentence is denoted in \textit{italics} and the ambiguous word is \textbf{bolded}. Note the dialectal use of the ``usted'' accusative form ``los''. Language denoted in parentheses.}
    \label{tab:data-examples}
    \vspace{-2mm}
\end{table*}

Neural machine translation (NMT) systems can produce high-quality, fluent output which are nearly indistinguishable from human translations, when evaluated at the sentence level.
This human-level parity has been shown to disappear, however, when evaluated in context \cite{laubli-etal-2018-machine, toral-etal-2018-attaining}.
This is unsurprising, because sentences are nearly always written by humans in some contextual setting, and are translated by translators in the same fashion.
Dismissing this context may create ambiguities that do not exist in the document as a whole, and in some cases, may make it impossible to correctly interpret the sentence.

Translation to another language must address ambiguities where the semantic or grammatical granularity of two sentences is imbalanced or mismatched.
Probably the most widely-known of these is grammatical gender, i.e.,\ when translating referential pronouns from a grammatically non-gendered language to a gendered one.
For example, when translating from English to French, the pronoun \emph{it} must be translated to \textit{il} or \textit{elle} depending on the grammatical gender of the antecedent noun, which may not be available in the same sentence.

The obvious path forward in addressing these issues is to move to contextual machine translation, in which sentences are no longer translated in isolation but with their source-side context.
Recent work has shown that transformers \cite{vaswani2017attention} are capable of handling longer sequences and improving performance on context-based evaluation \cite{sun-etal-2022-rethinking,post2023escaping}.
However, general contextual translation has a number of obstacles, foremost is the lack of available evaluation resources.  %
There are essentially two kinds of contextual evaluations: general metrics, which can theoretically be applied to any test set, and fixed test sets.
There is relatively little work in the former setting \cite{vernikos-etal-2022-embarrassingly,jiang-etal-2022-blonde}, and while they correlate with human judgments, they have not been proven capable of discriminating sentence-based from known-high-quality contextual systems.
For the latter, a number of high-quality evaluation sets exist \cite[Table~\ref{tab:thiswork}]{muller-etal-2018-large, lopes-etal-2020-document,bawden-etal-2018-evaluating,voita-etal-2019-good}, but they are limited both in language coverage and scope of phenomena.

In this work, we address this lack of evaluation data by extending coverage of existing datasets to more languages and contextual phenomena.
We:
\begin{itemize}
    \itemsep0em
    \item develop a pipeline that makes use of broad-language-coverage annotation tools and hand-developed rules to identify context-based phenomena in any test set;
    \item construct rules for five context-based phenomena (\S~\ref{sec:document-level-phenomena}) and seven language pairs (\S~\ref{sec:extraction-pipeline}): DE, ES, FR, IT, PL, PT, and RU with EN; and
    \item apply this toolchain to multiple datasets.
\end{itemize}
We show that this dataset, called \ctxpro, is capable of discriminating high-quality contextual systems from sentence-level ones.

\section{Contextual phenomena}
\label{sec:document-level-phenomena}

\begin{figure*}[ht]
    \centering
    \includegraphics[width=\textwidth]{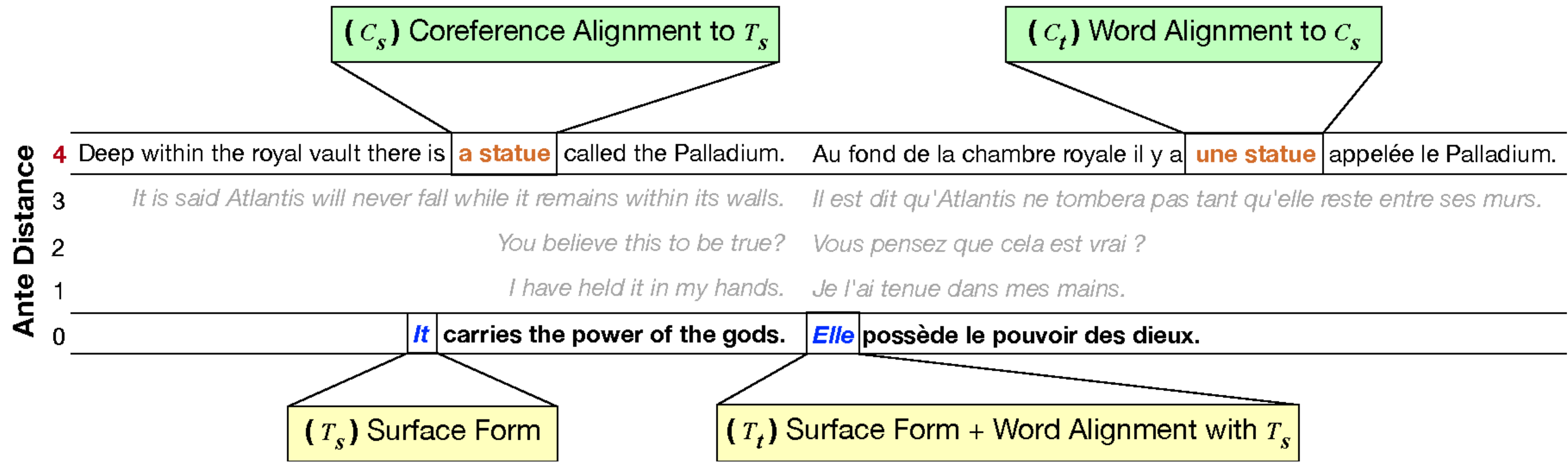}
    \caption{A diagram showing how the four key words for \gender identification are identified. The antecedent distance is determined by what sentence $C_e$ is found in. In order to be considered, $T_e$, $T_t$, $C_e$, $C_t$ would also have to pass morphological feature tests similar to those shown in Table \ref{tab:german-pronoun-rules}.}
    \label{fig:extraction}
    \vspace{-2mm}
\end{figure*}

A number of context-based phenomena which create ambiguities are common.
We display some examples in Table~\ref{tab:data-examples}.
Humans easily handle these ambiguities during translation, which nearly always takes place in context, so a machine translation system which ignores these issues will never reach human-level parity.
Some, such as lexical cohesion or fluency, are hard to quantify, while others, for example pronoun translation accuracy or word sense disambiguation, are easier.
These phenomena all present difficulties and even impossibilities to systems that translate sentences in isolation.
Our goal is to identify as many of these phenomena we can in a general way, such that we can create a general pipeline for isolating them, that can be reliably applied to any test set.

We describe each phenomena for comprehension and then provide our extraction methodology in order to identify when these ambiguities arise.

\subsection{Anaphoric pronouns}
\label{sec:anaphora}

Pronouns are a general descriptor that function as a placeholder for a noun phrase, providing the speaker with a more succinct form instead of repeatedly identifying an established referent.

In grammatical contexts, anaphora refers to the use of a pronoun to refer to a previously mentioned word or entity.
Pronouns for which the referent noun can be found in preceding contexts are called \textit{anaphora}; in contrast, \textit{cataphora} denotes situations where the referent noun follows the pronoun.
We do not consider cataphora in this paper.

\subsubsection{Gender}
\label{sec:gender}
Languages with gendered nouns require agreement with the appropriate gendered pronoun.
English, which makes no such distinction for inanimate objects, will use the pronoun ``it.''
In order to correctly translate ``it'' into Spanish, it is necessary to know what ``it'' refers to.
If ``it'' refers to a school, it would be translated differently (\textit{una} escuela) than if it refers to a heart (\textit{un} corazón).

Apart from a few exceptions, English does not make use of grammatical gender.
Machine translation often centers around translating either into or out of English with most of the paired languages expressing genders (masculine, feminine, and neuter), so there is a clear need to evaluate the translation of gender.
Further, removing English from the equation does not resolve the problem.
Gender assignment of inanimate objects is arbitrary which means that translating between two gendered languages is non-trivial.
In extreme cases, a language may exhibit \textit{``noun classes''} which behave similarly to gender, but may correlate more heavily with meaning.
A noun in Swahili is not grouped via an arbitrary \textit{gender} assignment, but is instead somewhat assigned to groups based on other labels such as \textit{animacy, items, plants}, or \textit{tools}.
These classes affect morphological agreement in ways that English does not express.
In any case, translating a pronoun that refers to a previously mentioned noun requires resolving this coreference in order to correctly generate the new pronoun.

\subsubsection{Animacy}
\label{sec:animacy}

Humans and animals are often treated differently grammatically than inanimate objects.
As stated, English makes no gender distinction for inanimate objects, though it does have gendered pronouns for \textit{animate} objects.
\emph{She} and \emph{he} are English pronouns used for humans and often animals but are rarely used to refer to inanimate objects.\footnote{A small exception occurs when inanimate objects are personified. A frequent example is boats, which are often referred to as \emph{she} in English.}
This results in an ambiguity when translating pronouns into English from languages that do not make this distinction.
For example, in English, \textit{she is in the kitchen} clearly refers to a person while \textit{it is in the kitchen} refers to a non-person.
In French, the word \textit{elle} would be used in both situations, requiring an MT system to make a choice.

\subsubsection{Formality}
\label{sec:formality}

Social expectations dictate language usage.
In many languages, this is explicitly lexicalized with different second-person pronouns and verb conjugations that distinguish intimate or familiar relationships from formal ones.
Examples include the \textit{tu/vous} distinction in French and \textit{du/Sie} in German.

Over time, English has lost its formal register in pronouns (often called the T-V distinction) which other languages frequently employ.
A common sentence ``Where are you?'' may have multiple interpretations determined by the addressee, but subtle cues in preceding context may indicate the level of formality or familiarity of the speaker---a ``sir'', the domain, or profession mentioned can clarify this.
When translating this sentence into French, the system must choose a register to produce either ``Où êtes-vous?'' or ``Où es-tu?''
There is often insufficient information to make the correct choice from just a single sentence.

\subsection{Verb Phrase Ellipsis}
\label{sec:vp-ellipsis}

Verb phrases can be dropped for emphasis, style, or brevity.
The manner in which they are ellipsed will follow the rules of syntax of the specific language.

\subsubsection{Isolated Auxiliaries}
\label{sec:auxiliary-replacement}

English auxiliaries (``do'', ``will'', ``would'') can occur as standalone verbs by taking the place of a verb phrase.
The question ``Will you walk with me?'' can be answered with a short ``I will.''
Many target languages require translation of the original head of the verb phrase rather than the modal or auxiliary.
Simply, ``I will'' must be translated as ``I will walk'' or rather ``I walk'' inflected in the future tense.
We limit this work to the aforementioned auxiliaries as they rarely have direct translations.

\begin{table*}[ht]
    \small
    \centering
    \begin{tabularx}{\textwidth}{l rrr rrr c rrr}
    \toprule
    &
    \multicolumn{3}{c}{English ($T_e$)} &
    \multicolumn{3}{c}{German ($T_t$)} &
    \multicolumn{1}{c}{Coref English ($C_e$)} &
    \multicolumn{3}{c}{Coref German ($C_t$) }
    \\
    \cmidrule(r){2-4} 
    \cmidrule(r){5-7}
    \cmidrule(r){8-8}
    \cmidrule(r){9-11}
    
    Rule &

    \multicolumn{1}{c}{Form} &
    \multicolumn{1}{c}{POS} &
    \multicolumn{1}{c}{Case} &

    \multicolumn{1}{c}{Form} &
    \multicolumn{1}{c}{POS} &
    \multicolumn{1}{c}{Case} &

    \multicolumn{1}{c}{POS} &

    \multicolumn{1}{c}{POS} &
    \multicolumn{1}{c}{Gender} &
    \multicolumn{1}{c}{Number} \\
    
    \midrule

\texttt{NOM.FEM.SING} & it & \texttt{PNOUN} & * & sie & \texttt{PNOUN} & Nom. & \texttt{NOUN} & \texttt{NOUN} & Fem. & Sing. \\
\texttt{NOM.MASC.SING} & it & \texttt{PNOUN} & * & er & \texttt{PNOUN} & Nom. & \texttt{NOUN} & \texttt{NOUN} & Masc. & Sing. \\
\texttt{NOM.NEUT.SING} & it & \texttt{PNOUN} & * & es & \texttt{PNOUN} & Nom. & \texttt{NOUN} & \texttt{NOUN} & Neut. & Sing. \\
\texttt{ACC.FEM.SING} & it & \texttt{PNOUN} & * & sie & \texttt{PNOUN} & Acc. & \texttt{NOUN} & \texttt{NOUN} & Fem. & Sing. \\
\texttt{ACC.MASC.SING} & it & \texttt{PNOUN} & * & ihn & \texttt{PNOUN} & Acc. & \texttt{NOUN} & \texttt{NOUN} & Masc. & Sing. \\
\texttt{ACC.NEUT.SING} & it & \texttt{PNOUN} & * & es & \texttt{PNOUN} & Acc. & \texttt{NOUN} & \texttt{NOUN} & Neut. & Sing. \\
\texttt{DAT.FEM.SING} & it & \texttt{PNOUN} & * & ihr & \texttt{PNOUN} & Dat. & \texttt{NOUN} & \texttt{NOUN} & Fem. & Sing. \\
\texttt{DAT.MASC.SING} & it & \texttt{PNOUN} & * & ihm & \texttt{PNOUN} & Dat. & \texttt{NOUN} & \texttt{NOUN} & Masc. & Sing. \\
\texttt{DAT.NEUT.SING} & it & \texttt{PNOUN} & * & ihm & \texttt{PNOUN} & Dat. & \texttt{NOUN} & \texttt{NOUN} & Neut. & Sing. \\
\texttt{NOM.INFORM.SING} & you & \texttt{PNOUN} & * & du & \texttt{PNOUN} & Nom. & - & - & - & - \\
\texttt{NOM.FORM+PLUR} & you & \texttt{PNOUN} & * & Sie & \texttt{PNOUN} & Nom. & - & - & - & - \\
\texttt{NOM.INFORM.PLUR} & you & \texttt{PNOUN} & * & ihr & \texttt{PNOUN} & Nom. & - & - & - & - \\
\texttt{ACC.INFORM.SING} & you & \texttt{PNOUN} & * & dich & \texttt{PNOUN} & Acc. & - & - & - & - \\
\texttt{ACC.FORM+PLUR} & you & \texttt{PNOUN} & * & Sie & \texttt{PNOUN} & Acc. & - & - & - & - \\
\texttt{ACC.INFORM.PLUR} & you & \texttt{PNOUN} & * & euch & \texttt{PNOUN} & Acc. & - & - & - & - \\
\texttt{DAT.INFORM.SING} & you & \texttt{PNOUN} & * & dir & \texttt{PNOUN} & Dat. & - & - & - & - \\
\texttt{DAT.FORM+PLUR} & you & \texttt{PNOUN} & * & ihnen & \texttt{PNOUN} & Dat. & - & - & - & - \\
\texttt{DAT.INFORM.PLUR} & you & \texttt{PNOUN} & * & euch & \texttt{PNOUN} & Dat. & - & - & - & - \\

    \bottomrule

    \end{tabularx}
    \caption{German criteria for all pronouns. 
    We expand from \citet{muller-etal-2018-large} to consider more cases (Accusative and Dative).
    English case is not used since the German annotations are more precise (English does not label Dative).
    \texttt{PNOUN} check in some cases is required to eliminate determiners (possessive adjectives instead of possessive pronouns)}
    \label{tab:german-pronoun-rules}
    \vspace{-2mm}
\end{table*}
\begin{table}[ht]
    \centering
    \begin{tabular}{l r r}
    \toprule
    &
    \multicolumn{1}{c}{English ($T_e$)} &
    \multicolumn{1}{c}{French ($T_t$)}
    \\
    \cmidrule(r){2-2} 
    \cmidrule(r){3-3}
    
    Rule &

    \multicolumn{1}{c}{Lemma} &

    \multicolumn{1}{c}{Illegal Lemmas}
    \\
    
    \midrule

    \texttt{DO.ELL}
    & do
    & faire, aller
    \\

    \texttt{WOULD.ELL}
    & would
    & faire, pouvoir
    \\

    \texttt{WILL.ELL}
    & will
    & aller, faire
    \\

    \bottomrule

    \end{tabular}
    \caption{French ellipsis Rules. English must have specified lemma. French alignment cannot have a lemma in the specified list.}
    \label{tab:french-rules}
    \vspace{-3mm}
\end{table}

\subsubsection{Inflection of Verb-less Nouns}
\label{sec:isolated-np}

Extreme ellipsis may remove entire portions of a sentence and render it a \textit{phrase}. 
English word order conveys grammatical role of nouns.
When elements of the original sentence, such as the verb, are ellipsed, it may be impossible to infer the grammatical case of any remaining nouns which have no inflection.
Translation into languages with case systems suffers.
\citet{voita-etal-2019-good} exemplifies using the phrase: ``You call her your friend but have you been to
her home? Her work?''
To translate this phrase into Russian, it is necessary to know that ``her work'' has the same grammatical case as ``her home'' in the previous sentence.

\section{Extraction Pipeline}
\label{sec:extraction-pipeline}

\begin{table*}[t]
    \centering
\begin{tabular}{l rrrrrrr}
\hline
 & DE & FR & RU & PL & PT & IT & ES \\ \hline
\gender & 147k & 291k & 113k & 117k & 127k & 36k & 96k \\
\animacy* & 80k & 145k & 66k & 39k & 38k & 20k & 84k \\
\formality & 3.9M & 5.7M & 3.6M & 1.7M & 857k & 833k & 10.1M \\
\auxiliary & 4414 & 27.6k & 39.1k & 34.2k & 30.2k & 17.5k & 29.6k \\
\inflection & - & - & 2.6M & 3.2M & - & - & - \\ \hline
\textsc{\# lines} & 22.5M & 41.9M & 25.9M & 77.2M & 33.2M & 35.2M & 61.4M \\
\textsc{\% Extracted} & 18\%  & 14\% & 25\% & 6.6\% & 3.1\% & 2.5\% & 16.7\% \\
\textsc{\%-Coreference} & 0.7\% & 0.8\% & 0.6\% & 0.2\% & 0.5\% & 0.2\% &  0.2\% \\ \hline
\end{tabular}
    \caption{OpenSubtitles2018 Extraction Statistics for each category. \textsc{\# lines} indicates the total number of lines in OpenSubtitles for the EN-XX language pair. \textsc{\% Extracted} indicates the percent of the dataset that was extracted. \textsc{\%-Coreference}
    indicates the classes that require a strict antecedent (\gender and \auxiliary). (*) \animacy was created by reversing a subset of the \gender class so it is not used to calculate \textsc{Extracted} because of the overlap.}
    \label{tab:opensubs-stats}
    \vspace{-2mm}
\end{table*}

Our pipeline functions by identifying up to four key tokens and ensuring each token matches a set of predefined criteria.
The four components are: (1) The source (English) token defined as $T_s$, the target (non-English) token defined as $T_t$, the source token which conveys the contextual information required to resolve the ambiguity defined as $C_s$, and the target token aligned to $C_s$ defined as $C_t$.
These relationships are illustrated in Figure \ref{fig:extraction}.
Contextual information is defined by a contextual relationship, $Q$, which has an associated solver.
The predefined criteria is a set of rules, $R$.

We can identify ambiguous sentences by:
\begin{enumerate}
    \itemsep0em 
    \item For each source–target sentence pair, apply word alignment. Each aligned pair of words forms a potential $T_s$–$T_t$ pair.
    \item Ensure $T_s$ meets all criteria $R_{T_s}$
    \item Ensure $T_t$ meets all criteria $R_{T_t}$
    \item Apply a solver for the contextual relationship, $Q$ to the English token $T_s$ and its preceding context to identify $C_s$.
    \item Ensure $C_s$ meets all criteria $R_{C_s}$.
    \item Identify the target token $C_t$ via word alignment to $C_s$. If translation conveys semantic symmetry, this token \textit{also} has a contextual relationship with $T_t$.
    \item Ensure $C_t$ meets all criteria $R_{C_t}$
\end{enumerate}

Consider the ambiguity of pronoun resolution.
\citet{muller-etal-2018-large} first proposed a pipeline for extracting ambiguous translations of English ``it'' to German nominatives (``er'', ``es'', and ``sie'').
We can explain their methodology\footnote{\citet{muller-etal-2018-large} performs an extra coreference check on the target side that we do not.} via the aforementioned definition.
The following identifies all ambiguities where the English ``it'' is translated as ``sie.''
\begin{enumerate}
    \itemsep0em 
    \item For each source-target sentence pair, apply word alignment. Each aligned pair of words forms a potential $T_s$–$T_t$ pair.
    \item Ensure $T_s$ is the word ``it''
    \item Ensure $T_t$ is the word ``sie''
    \item The contextual information to resolve the ambiguity is its antecedent---expressed via a coreference relationship. Apply a coreference resolver ($Q$) to identify $C_s$.
    \item Ensure $C_s$ is a noun (not another pronoun).
    \item Identify $C_t$ via word alignment.
    \item Ensure $C_t$ is a feminine, singular noun.
\end{enumerate}
The same criteria could be enumerated for the masculine and neuter equivalents, appropriately changing gender and surface form checks.

To extract a specific phenomenon and language, a ``rule'' ($R$) must be written which specifies features that $T_s$, $T_t$, $C_s$, and $C_t$ must have.
These features can range from part-of-speech, lemma, gender, case, plurality or others.
The manner in which these four components are identified creates the adaptability for each phenomena.

\paragraph{Gender}  Following previous works, we retrieve $T_s$ and $T_t$ based on surface form and word alignment.
$C_s$ is a noun discovered via coreference chain.
If the coreference is a noun phrase, the head of the phrase is used.
$C_t$ is retrieved via word alignment.
$C_t$ must match the same morphological features present in $T_t$ (e.g., gender and number).

\paragraph{Animacy} As explained in Section \ref{sec:animacy}, the animacy ambiguity that we consider occurs when translating from the gendered languages \textit{into} English (whereas the gender ambiguity occurs when translating \textit{out-of} English).
To extract these examples, we use the same rules as \gender, but we reverse the language direction for inference.

\paragraph{Formality}
The distinction of formality is the lack of a consistent or discrete $C_s$ which informs the level of formality.
Translating between English and a T-V language is always ambiguous with respect to the second person so we forgo using a contextual resolver $Q$ to identify the appropriate context.

\paragraph{Auxiliary} $T_s$ is extracted from a pre-constructed list of auxiliaries---similar to those mentioned in Section \ref{sec:auxiliary-replacement}.
$T_t$, identified via word alignment, cannot occur in a pre-constructed list of forbidden translations.
These translations are meant to prevent valid translations of auxiliaries, rather than the ambiguous ellipsed forms.
For example, ``to do'' translated as a form of ``faire'' in French, is a direct translation, and is likely not representative of an ellipsed form.
Contrarily, ``to do'' translated as a form of ``savoir'' in French is not a direct translation and is indicative of a previous occurrence of English ``to know.''
$C_t$ can be identified by finding the most recent occurrence of the same verb $T_t$, and $C_s$ is retrieved from word alignment with $C_t$.

\paragraph{Inflection} $T_s$ and $T_t$ can be of any form and any case.
Any aligned noun pair ($T_s$ and $T_t$) that occurs without an accompanying verb is ambiguous.
$C_t$ is identified as the most recent occurrence of \textit{any noun} occurring in the same case as $T_t$.
We assume the verb phrase surrounding $C_t$ was ellipsed when generating $T_t$.
We align $C_t$ to find $C_s$.

We use FastCoref \cite{otmazgin2022fcoref} to perform English coreference resolution, simalign \cite{jalili-sabet-etal-2020-simalign} to perform cross-lingual word alignment, and SpaCy\footnote{\url{https://spacy.io/usage/models\#languages}} to extract all other morphological features.
We provide a larger list of our criteria in Appendix \ref{sec:materials-appendix}.

\subsection{Application to OpenSubtitles}

We apply our extractor to the OpenSubtitles2018 dataset \cite{lison-tiedemann-2016-opensubtitles2016} following previous work \citep{muller-etal-2018-large, lopes-etal-2020-document}.
It comprises conversational dialog extracted from film and television subtitles.
The conversational nature means plenty of context-based phenomenon occur.
In Table \ref{tab:opensubs-stats}, we present the total number of instances we extracted from Open Subtitles.

The fraction of the dataset that contains the phenomenon we target varies from language to language.
This stems from the number of forms in each language, the number of genders, as well as translation standards.
German, for instance, has very few \auxiliary examples.
We speculate this is due to German having similar auxiliary features as English so many examples were filtered out due to our ``forbidden translation'' criteria.

Some categories are extremely common.
\formality is invoked every time the second-person is used, which is frequent in conversational speech.
\inflection also has high occurrences since there was relatively little filtering on the extracted examples.
\gender and \auxiliary are \textit{very rarely} extracted---less than 1\% of the time in all languages.
A 1\% error rate is extreme when deploying at scale.
Further, test sets, in nature, are small.
If only 1\% of the test set challenges contextual models, the results may be insignificant.

To form the dev, devtest, and test splits, we apply the following approach.
For each label within a category, we ensure there are at least 100 examples.
If there are fewer, we keep all examples for test.
If there are more, we split the most recent years of OpenSubtitles into a 1:1:5 ratio for dev:devtest:test, limiting the test set's maximum size to 5000 examples per label.
One label is roughly one surface form, but corresponds to one ``rule'' (a set of criteria $R$) or one row as shown in Table \ref{tab:german-pronoun-rules}.

\section{Quantitative Evaluation}
\label{sec:evaluation}

\begin{table*}[ht]
        \small
        \centering
\setlength{\tabcolsep}{3pt}
    \begin{tabular}{ll rrrrrrr  rrrrrrr}

    \toprule
    & & \multicolumn{7}{c}{Generative Accuracy (\%)} & \multicolumn{7}{c}{COMET} \\
    \cmidrule(lr){3-9} \cmidrule(lr){10-16}
    & 
    & \multicolumn{1}{c}{DE}
    & \multicolumn{1}{c}{ES}
    & \multicolumn{1}{c}{FR}
    & \multicolumn{1}{c}{IT}
    & \multicolumn{1}{c}{PL}
    & \multicolumn{1}{c}{PT}
    & \multicolumn{1}{c}{RU} 

    & \multicolumn{1}{c}{DE}
    & \multicolumn{1}{c}{ES}
    & \multicolumn{1}{c}{FR}
    & \multicolumn{1}{c}{IT}
    & \multicolumn{1}{c}{PL}
    & \multicolumn{1}{c}{PT}
    & \multicolumn{1}{c}{RU} 
    \\
    \cmidrule{1-16}

    \multirow{3}{*}{\gender} & sent. 
    
    & 48.1 & 34.6 & 40.2 & 51.1 & 32.8 & 44.3 & 35.9 %
    
    & 0.23 & 0.50 & 0.33 & 0.43 & 0.51 & 0.52 & 0.36 %
    
    \\

    & doc. 
    & \textbf{73.3} & \textbf{47.4} & \textbf{59.0} & \textbf{68.3} & \textbf{50.2} & \textbf{64.3} & \textbf{51.8} %
    & \textbf{0.31} & \textbf{0.52} & \textbf{0.43} & \textbf{0.48} & \textbf{0.54} & \textbf{0.57} & \textbf{0.42} %
    
    \\

    & 
    & +25.2 & +12.8 & +18.8 & +17.2 & +17.4 & +20.0 & +15.9 %
    &  +0.08 & +0.02 & +0.09 & +0.05 & +0.03 & +0.05 & +0.06 %
    \\

    \midrule

    \multirow{3}{*}{\animacy} & sent.
    & 61.0 & 84.4 & 68.0 & 81.4 & 57.6 & 64.1 & 55.4 %
    & 0.27 & 0.53 & 0.40 & 0.42 & 0.25 & 0.43 & 0.19 %
    \\
    & doc. 
    & \textbf{74.1} & \textbf{87.8} & \textbf{75.2} & \textbf{86.1} & \textbf{70.5} & \textbf{79.5} & \textbf{71.6} %
    & \textbf{0.38} & \textbf{0.58} & \textbf{0.49} & \textbf{0.46} & \textbf{0.31} & \textbf{0.55} & \textbf{0.34} %
    \\
    
    & 
    & +13.1 & +3.4 & +7.2 & +4.7 & +12.9 & +15.4 & +16.2 %
    & +0.11 & +0.05 & +0.09 & +0.04 & +0.06 & +0.12 & +0.15 %
    \\

    \midrule

    \multirow{3}{*}{\formality} & sent.
    & 44.0 & 31.7 & 40.6 & 38.9 & 25.3 & 40.1 & 55.4 %
    & \textbf{0.32} & 0.54 & 0.45 & 0.47 & \textbf{0.51} & \textbf{0.59} & 0.57 %
    \\
    
    & doc. 
    & \textbf{53.6} & \textbf{35.9} & \textbf{51.5} & \textbf{46.1} & \textbf{31.6} & \textbf{47.2} & \textbf{62.5} %
    & \textbf{0.32} & \textbf{0.55} & \textbf{0.48} & \textbf{0.48} & \textbf{0.51} & \textbf{0.59} & \textbf{0.58} %
    \\
    & 
    & +9.6 & +4.2 & +10.9 & +7.2 & +6.3 & +7.1 & +7.1 %
    & +0.0 & +0.01 & +0.03 & +0.01 & +0.0 & +0.0 & + 0.01 %
    \\

    \midrule
    \multirow{3}{*}{\auxiliary} & sent.
    & 7.8 & 3.3 & 1.3 & 4.0 & 8.2 & 9.2 & 5.7 %
    & -0.27 & -0.06 & -0.34 & -0.02 & 0.10 & 0.03 & -0.09 %
    \\
    & doc. 
    & \textbf{40.0} & \textbf{52.0} & \textbf{32.2} & \textbf{40.7} & \textbf{49.9} & \textbf{53.8} & \textbf{49.0} %
    & \textbf{0.04} & \textbf{0.54} & \textbf{0.20} & \textbf{0.38} & \textbf{0.53} & \textbf{0.60} & \textbf{0.49} %
    \\
    & 
    & +32.2 & +48.7 & +30.9 & +36.7 & +41.7 & +44.6 & +43.3 %
    & +0.31 & +0.60 & +0.54 & +0.40 & +0.43 & +0.57 & +0.58 %
    \\

    \midrule

    \multirow{3}{*}{\inflection} & sent.
    & - & - & - & - & 41.3 & - & 34.6 %
    & - & - & - & - & 0.57 & - & 0.47 %
    \\
    & doc. 
    & - & - & - & - & \textbf{53.2} & - & \textbf{48.3} %
    & - & - & - & - & \textbf{0.68} & - & \textbf{0.56} %
    \\
    & 
    & - & - & - & - & +11.9 & - & +13.7 %
    & - & - & - & - & +0.11 & - & +0.09 %
    \\
    \bottomrule

    \end{tabular}
    \caption{Generative evaluation percent accuracy scores (left section) evaluation ability to produce expected form; COMET scores (right section) evaluate the translation quality of this model; \emph{sent.} denotes that no additional context was given while \emph{doc.} was given five consecutive sentences for context. All translations made using DeepL commercial API. \animacy is \textit{into} English. All others are \textit{out of} English}
    \label{tab:results}
    \vspace{-2mm}

\end{table*}

Our goal is to show that our test sets can usefully discriminate between sentence-level and context-aware systems.
An impediment to this goal is the lack of contextual machine translation models across languages for use in comparison and evaluation, and the difficulty in building them.
Consequently, we turn to a commercial system, DeepL, which is alone among commercial providers in advertising contextual translation.\footnote{\url{https://www.deepl.com/docs-api/general/working-with-context}}
We translate with document-context by providing DeepL with context when translating, and compare to the same model translating without context at the sentence level.
We show that a contextual system appropriately benefits from the additional context and gains significance performance on this test set.

Many works release their evaluation sets with the assumption of contrastive evaluation \cite{muller-etal-2018-large,lopes-etal-2020-document,voita-etal-2019-good}, where the test is whether a model assigns a higher score to correct data than to linguistically-manipulated counterparts.
This assumption ignores the fact that machine translation is a generative problem and should be evaluated as such.
Recent work \cite{post2023escaping} confronts this problem and proposes generative evaluation as an alternative, showing a wide gap between contextual and sentence-level systems that is only observed under generative evaluation.
Translations are counted as correct if the \textit{expected surface form} is present anywhere in the model's output---matching the entire word and not simply a substring. 
We follow this approach in our evaluation.

We validate our data by showing it (1) adequately addresses context-based phenomena and (2) is sufficiently challenging.
We demonstrate the former by showing that a context-aware translation model consistently outperforms a context-less equivalent.
We see the latter is true as the contextual model does not solve the problem.
There is still significant context-aware work to be done.

\subsection{Accuracy}

\matt{You need to introduce how you test more gently.}

We begin by translating sentences both with and without context, using at most five sentences of context.
To limit API calls, we run a subsample of our produced evaluation sets.
We limit each category (\gender, \animacy, \formality, \auxiliary, and \inflection) to approximately 10k total examples, divided evenly amongst the categories labels.
To extract the final sentence for scoring purposes, we apply segmentation using the \ersatz segmenter \cite{wicks-post-2022-sentence}.

The results in Table \ref{tab:results} clearly show that the DeepL model with additional context \textit{far outperforms} its sentence-level equivalent.\footnote{Ideally we would make the same comparison between document- and sentence-level translation with other commercial systems, but there is no way to prevent them from applying sentence-level segmentation to the document-context string.}
Many of these evaluation examples have specific preceding context that needs to be used in order to correctly translate the ambiguity.
\formality is a slight exception.
There is little to no guarantee that explicit cues are given to convey the nature of the relationship between the speaker and addressee, yet preceding context still benefits an average of 9 percentage points across all languages.
\auxiliary is a task of translating verbs.
A random guess would equate to sampling from the distribution of verbs in a language--which results in low success rates.
Translating \auxiliary with context increases from nearly never correct to a roughly 50\% accuracy rate. 
Translating \animacy has higher sentence-level baselines than some of the other categories.
We attribute this to other semantic cues towards \animacy which are less arbitrary than something such as \gender assignment.
For instance, if a noun \emph{talks}, it is likely animate, while a noun that \emph{is thrown} is likely inanimate.
Similarly, \inflection may have some sentence-internal cues.
Certain nouns may have a majority class, or preceding prepositions ((``with'', ``for'', ``in'', etc.) may indicate case.
This is similar to the intrasentential coreference found with pronouns, which makes some occurrences easier than others.
Nonetheless, additional context aids the model.
In every category, the context-aware model shows consistent gains over its context-less variant.

\subsection{Automatic metric}

\matt{BLEU scores could go to an appendix.}
\matt{This section's rewrite/update awaiting COMET20 scores.}
We also present COMET scores \cite{rei-etal-2020-comet} in Table \ref{tab:results}.
Across all categories and language pairs, COMET shows improvement when the system leverages additional context.
The consistent improvement in COMET reinforces the trends we see with the generative evaluation metric.
The one exception is the \formality class which has minimal differences between the sentence-level and contextual inputs.
COMET rewards synonyms and we suspect formal and informal surface forms have more similar encodings in COMET models than these other grammatical forms.
A surface-based metric would better capture the gains that can be seen from the accuracy scores, which is indeed what we find (Table~\ref{tab:bleu-alltables} in Appendix~\ref{sec:materials-appendix}).

\begin{table*}[ht]
    \centering

\begin{tabular}{lp{6.5cm}p{6.5cm}}

\toprule
\multicolumn{1}{c}{Error Type}
 & \multicolumn{1}{c}{English} & \multicolumn{1}{c}{French} \\
 \midrule
 
\multicolumn{1}{l}{\multirow{5}{*}{Coreference}} &
We got any ideas what these \underline{\textit{\textbf{guys}}} were after?
& 
Une idée de ce que voulaient ces \underline{\textbf{\textit{gars}}}?
\\

\multicolumn{1}{l}{} &
No, CEO is on his way down to talk to us now. &
Non, le PDG arrive pour nous le dire.
\\

\multicolumn{1}{l}{} &
So far, everyone we've talked to hasn't really given us much. &
Tous ceux à qui on a parlés ne nous ont rien appris.
\\

\multicolumn{1}{l}{} &
Makes sense. & 
C'est logique.
\\

\multicolumn{1}{l}{} &
\underline{\textit{\textbf{They}}} don't want us to know what they're working on here.
 & 
\underline{\textit{\textbf{Ils}}} ne veulent pas qu'on sache ce qu'ils font.
\\

\midrule

\multicolumn{1}{l}{\multirow{2}{*}{Alignment}} &
As you know the \underline{\textbf{\textit{discipline}}} of media espionage is a new one.
&
Comme vous le savez, l'\underline{\textbf{\textit{espionnage}}} médiatique est une nouvelle discipline.
\\

\multicolumn{1}{l}{} &
Oh yes, \underline{\textit{\textbf{it}}} is everywhere. &
\underline{\textit{\textbf{Il}}} est partout.
\\
\midrule

\multicolumn{1}{l}{\multirow{2}{*}{Translation}} &
You know more about \underline{\textit{\textbf{signatures}}} than most of \underline{\textbf{\textit{them}}} put together.
&
Vous en savez plur sur ces \underline{\textbf{\textit{tueurs}}} (\texttt{en: killers}) qu'\textbf{\textit{\underline{eux}}} tous réunis
\\

 \bottomrule
\end{tabular}
    \caption{In a sample of 100 extracted items, 8 errors were found. This table shows 3 of these errors made by the extraction pipeline on the French Gender set. The \underline{\textbf{\textit{indicated}}} words show the pronouns in French and English, as well as their antecedents. Some examples fit into multiple categories, but these show the most evident error type. \texttt{en:} indicates the English translation of French word.}
    \label{tab:french-errors}
\end{table*}

\section{Qualitative Evaluation}
\label{sec:quality}

Our extraction pipeline relies on handbuilt rules applied to the outputs of automatic tools.
As a result, the process is noisy and may be susceptible to errors.
The previous section showed that a contextual system does better on our test sets than its sentence-based counterpart, and there is no reason we can think of to suspect that errors would systematically benefit the contextual system.
However, in the interest of completeness, we took a more qualitative look at the data.
This includes a systematic manual review (\S~\ref{section:manual}), direct comparison with prior work (\S~\ref{section:cmp}), and an error analysis (\S~\ref{section:errors}).

\subsection{Manual review}
\label{section:manual}

Previous work in automatic test set production has not typically included a manual analysis of rule quality.
To build confidence in these automatic extraction methodologies, we sampled 100 random test examples from the extracted English–French \gender set and manually reviewed and annotated them for errors.
We find that 92 of the extracted examples are correct.
Three more were questionably incorrect---with correct translations and alignments---yet had atypical coreference resolutions that were difficult for our human reviewer to understand. 
Of the remaining five, two had a non-referential pronoun.
One such example ``What is it?'' was used in the sense of ``What's wrong?'' rather than ``What is that?''
In the former, ``it'' has no valid antecedent, yet it was extracted.

We present the remaining three errors in Table \ref{tab:french-errors}, where they demonstrate where errors arise at each step in the pipeline.
The Coreference Error is a clear mistake.
``They don't want us to know what they're working on'' refers to the people being talked to, and not ``these guys''---who instead seemed to be criminals who broke into a company.
The Alignment Error is an unfortunate combination of a bad alignment and inconsistent translation.
``the discipline'' is aligned to the word ``espionnage.''
``discipline'' in French is a feminine noun, while ``espionnage'' is masculine.
The French ``il'' is masculine, and thus has ``espionnage'' as an antecedent despite the English having ``the discipline.''
This coincidental error caused this example to still be extracted.
Lastly, one of these examples seemed to have a typo in the English transcript.
The word ``signatures'' seemed to be incorrect.
We suspect the correct transcription word was ``serial killers.''
Given the inconsistent context on the English side, we suspect the neural coreference model had difficulties resolving this.

\subsection{Comparison with prior work}
\label{section:cmp}

Since our extraction framework is largely based on that of \citet{muller-etal-2018-large}, we expect to have a similar quality of extracted rules (or better, since the underlying annotations tools have improved).
We thus undertake a comparison to the data that they released.
When applying our pipeline to the German–English OpenSubtitles data, we extract 147,211 sentences that have ambiguous pronoun usage.
M\"{u}ller did not report their raw extraction numbers, but their release includes 12,000 examples, balanced across gender (but not distance).
We therefore focus our analysis on this subset.

Since their pipeline contained a target-side coreference check that we do not have, one might think their pipeline would be a stricter selection process, but we find the opposite to be true.
Our pipeline's selection overlaps with only half of ContraPro (6,003 sentences), rejecting the other half (5,997 sentences).
An analysis of this rejected portion of ContraPro turns up some explanations.
ContraPro extracts three categories of German pronouns corresponding to neuter, masculine, and feminine genders.
For \emph{er} and \emph{sie}, we rejected roughly 25\% of the ContraPro examples; however, we rejected over 75\% of the neuter examples from ContraPro.
Upon review, we found a substantial number of non-referential instances.
These examples include sentences such as ``It was your duty.'', ``It would have been all right if it wasn't for you.'' and ``It was one of those California Spanish houses'' that all have either a non-specified referent or have a passive construction.
The inclusion of these examples points to inaccurate coference chains, likely explained by their use of older corefence tools.

Our extraction employs strict criteria to find the head of a span during coreference and alignment.
The head is used for the gender, person, and number checks included in the definition of $R$ (\S~\ref{sec:extraction-pipeline}).
From our understanding of Müller's work, they did not include this check.
Mistakes are inherent to any automatic process, and likely persist in our dataset as well.
Our analysis here lends some confidence to the belief that tighter selection criteria and improved underlying tools result in better data.

\subsection{Model analysis}
\label{section:errors}

Absent sufficient information, the translation of ambiguous words will regress to their proportions in the training data.
For pronouns, this would be the neuter or masculine class; for auxiliaries, the direct translation (the ``Illegal Lemma'' in $R$).

We examine the English–German model outputs.
Our evaluation sets have balanced counts across genders, so a correct model would produce a neuter pronoun roughly one-third of the time.
Instead, this sentence-level model produces either ``es'' or ``ihm'' (the German neuter pronouns) closer to two-thirds of the time.
This contextual model has better performance producing the neuter pronouns about 40\% of the time.
This problem is well-known, but other issues are not as well documented.

The auxiliary category had the worst scores, both in terms of how low the sentence-level model was performing as well as the absolute increase from adding context.
The cause of these scores becomes obvious as we examine the model outputs.
To generate the rules for the \auxiliary class, we enumerated illegal lemmas that represent the most common direct translations of English modals as described in Section \ref{sec:extraction-pipeline}.
Ideally, a model would never generate these verbs for our evaluation set unless part of a larger verb phrase construction.
We find the sentence-level model generates a translation that contains a form of one of these lemmas approximately two-thirds of the time.
Conversely, the contextual model generates these closer to one-third of the time.

\begin{table}[h]
    \centering
\begin{tabular}{l rrr}
\toprule
 & \multicolumn{1}{c}{DE} & RU & PL \\
 \cmidrule(lr){2-4}
\multicolumn{1}{l}{\gender} & 135 & 64 & 13 \\
\multicolumn{1}{l}{\formality} & 540 & 416 & 4  \\
\multicolumn{1}{l}{\auxiliary} & 1 & 0 & 0  \\
\multicolumn{1}{l}{\inflection} & - & 14 & 1 \\
\midrule
\multicolumn{1}{l}{\textsc{WMT \# lines}} & 6454 & 7038 & 1000 \\
\bottomrule
\end{tabular}
    \caption{Counts on the number of extracted examples from WMT 2019-2022 (when available) test sets.}
    \label{tab:wmt-stats}
    \vspace{-2mm}
\end{table}

\section{Analysis of WMT test sets}

As previously earlier, this pipeline is easily applied to new data and test sets.
We demonstrate this by applying it to the 2019--2022 WMT newswire test sets \cite{barrault-etal-2019-findings,barrault-etal-2020-findings,akhbardeh-etal-2021-findings,kocmi-etal-2022-findings}.
In so doing, we find phenomena in a similar proportion of sentences to OpenSubtitles, but with a different distribution;
there is a higher rate of \gender but smaller of \formality and \auxiliary.
In Table \ref{tab:wmt-stats}, we present the total number of examples discovered in WMT 2019-2022 in \texttt{en-de}, \texttt{en-ru}, and \texttt{en-pl} (when available).
The newswire text hardly ever contains the \auxiliary type of ambiguity.
Formality comprises the bulk of the examples, and upon further inspection, we find a severe bias towards the formal register, with a 1 to 7 ratio of informal to formal---likely due to the characteristics of the domain.
Further, we suspect the sparseness in contextual ambiguities is important to consider when evaluating these systems.

\section{Related Work}
\label{sec:related-works}

Work in contextual machine translation can be divided into three categories: (1) the publication of resources, similar to this work; (2) alterations on the training paradigm via architecture or data input; (3) evaluation metrics.

This work largely follows the path set forth by those who have previously published resources on the detection of gender, pronouns, and formality \cite{guillou-hardmeier-2016-protest, muller-etal-2018-large, bawden-etal-2018-evaluating, voita-etal-2019-good, lopes-etal-2020-document}.
\cite{currey-etal-2022-mt} produces a gender-based evaluation dataset using human annotators, but covers the complement of this work: gender assigned to humans rather than inanimate objects.
In addition to the manual pipelines, recent work has been done to promote the automatic detection of these phenomena.
\citet{nadejde-etal-2022-cocoa} implements a cross-lingual mutual information metric that tags words as needing additional context.
The tags were found to often overlap with the variety discussed in this work.
\citet{fernandes-etal-2023-translation} also use a mutual-information based score to select data that is then used to derive a similar rule-based extraction approach, but do not release evaluation sets.

A substantial amount of work has been done to allow traditional neural models to handle additional input.
Some approaches involve more complex architectures or modifications to training paradigms incorporate longer sequences \cite{miculicich-etal-2018-document,bao-etal-2021-g},
but \citet{sun-etal-2022-rethinking} showed that unaltered Transformers can handle longer sequences.
Other work has focused on leveraging and cleaning the available data, since large-scale document bitext is lacking \cite{junczysdowmunt2019microsoft,post2023escaping}.

Lastly, many have realized that BLEU, COMET, or other sentence-level metrics will not address the distinction in document-level performance.
\citet{vernikos-etal-2022-embarrassingly} proposed a new method for adjusting current methods to adjust for document-level inputs.
\citet{jiang-etal-2022-blonde} proposed BlonDe, an entirely novel metric for document-level evaluation.
We hope this work complements these works and serves to further the field in its aspirations towards true context-aware translation.

\section{Summary}

Machine translation systems face a performance ceiling that can't be overcome so long as they continue to operate at the sentence level.
A major obstacle to that transition is the unavailability of test sets in many languages and for many contextual phenomena.
The goal of this work has been to help address that problem.
The extraction pipeline proposed in this paper can be used to identify and generate new test sets which contain linguistic phenomena that can only be consistently translated by contextual systems.
The application of our pipeline to the OpenSubtitles dataset in seven languages provides a new set of evaluation sets including a wider set of languages and phenomena than were available before.
Further, we hope that the extensibility of our pipeline to new phenomena and languages allows for others to build upon this work to expand resources and coverage.
The \ctxpro datasets and extraction pipeline are available as open source from \url{https://github.com/rewicks/ctxpro}.

\bibliography{anthology,custom}
\bibliographystyle{acl_natbib}

\appendix

\section{Additional Materials}
\label{sec:materials-appendix}

\begin{table}[h!]
    \centering
    \begin{tabularx}{\columnwidth}{l r p{3cm}}
    \toprule
    &
    \multicolumn{1}{c}{English ($T_e$)} &
    \multicolumn{1}{c}{German ($T_t$)}
    \\
    \cmidrule(r){2-2} 
    \cmidrule(r){3-3}
    
    Rule &

    \multicolumn{1}{c}{Lemma} &

    \multicolumn{1}{c}{Illegal Lemmas}
    \\
    
    \midrule

    \texttt{DO.ELL}
    & do
    & machen, tun, haben, können
    \\
    
    \texttt{WOULD.ELL}
    & would
    & machen, tun, haben
    \\

    \texttt{WILL.ELL}
    & will
    & machen, tun, haben, werden
    \\

    \bottomrule

    \end{tabularx}
    \caption{German auxiliary rules. English must have specified lemma. German alignment cannot have a lemma in the specified list.}
    \label{tab:german-auxiliary-rules}
\end{table}

\begin{table}[h!]
    \centering
    \begin{tabularx}{\columnwidth}{l r p{3cm}}
    \toprule
    &
    \multicolumn{1}{c}{English ($T_e$)} &
    \multicolumn{1}{c}{Polish ($T_t$)}
    \\
    \cmidrule(r){2-2} 
    \cmidrule(r){3-3}
    
    Rule &

    \multicolumn{1}{c}{Lemma} &

    \multicolumn{1}{c}{Illegal Lemmas}
    \\
    
    \midrule

    \texttt{DO.ELL}
    & do
    & robić
    \\
    
    \texttt{WOULD.ELL}
    & would
    & robić, by być, być, by, móc
    \\

    \texttt{WILL.ELL}
    & will
    & robić, by być, być, by, móc, iść
    \\

    \bottomrule

    \end{tabularx}
    \caption{Polish auxiliary rules. English must have specified lemma. Polish alignment cannot have a lemma in the specified list.}
    \label{tab:polish-aux-rules}
\end{table}

\begin{table}[h!]
    \centering
    \begin{tabular}{l r r}
    \toprule
    &
    \multicolumn{1}{c}{English ($T_e$)} &
    \multicolumn{1}{c}{Russian ($T_t$)}
    \\
    \cmidrule(r){2-2} 
    \cmidrule(r){3-3}
    
    Rule &

    \multicolumn{1}{c}{Lemma} &

    \multicolumn{1}{c}{Illegal Lemmas}
    \\
    
    \midrule

    \texttt{DO.ELL}
    & do
    & \foreignlanguage{russian}{Делать}
    \\
    
    \texttt{WOULD.ELL}
    & would
    & \foreignlanguage{russian}{Делать}
    \\

    \texttt{WILL.ELL}
    & will
    & \foreignlanguage{russian}{Делать}
    \\

    \bottomrule

    \end{tabular}
    \caption{Russian auxiliary rules. English must have specified lemma. Russian alignment cannot have a lemma in the specified list.}
    \label{tab:russian-aux-rules}
\end{table}

\begin{table}[h!]
    \centering
    \begin{tabularx}{\columnwidth}{l r p{3cm}}
    \toprule
    &
    \multicolumn{1}{c}{English ($T_e$)} &
    \multicolumn{1}{c}{Portugese ($T_t$)}
    \\
    \cmidrule(r){2-2} 
    \cmidrule(r){3-3}
    
    Rule &

    \multicolumn{1}{c}{Lemma} &

    \multicolumn{1}{c}{Illegal Lemmas}
    \\
    
    \midrule

    \texttt{DO.ELL}
    & do
    & fazer
    \\
    
    \texttt{WOULD.ELL}
    & would
    & fazer, poder
    \\

    \texttt{WILL.ELL}
    & will
    & fazer, ir
    \\

    \bottomrule

    \end{tabularx}
    \caption{Portuguese auxiliary rules. English must have specified lemma. Portuguese alignment cannot have a lemma in the specified list.}
    \label{tab:polish-aux-rules}
\end{table}

\begin{table}[ht!]
    \centering
    \begin{tabularx}{\columnwidth}{l r p{3cm}}
    \toprule
    &
    \multicolumn{1}{c}{English ($T_e$)} &
    \multicolumn{1}{c}{Italian ($T_t$)}
    \\
    \cmidrule(r){2-2} 
    \cmidrule(r){3-3}
    
    Rule &

    \multicolumn{1}{c}{Lemma} &

    \multicolumn{1}{c}{Illegal Lemmas}
    \\
    
    \midrule

    \texttt{DO.ELL}
    & do
    & fare
    \\
    
    \texttt{WOULD.ELL}
    & would
    & fare, potere, volere
    \\

    \texttt{WILL.ELL}
    & will
    & fare, andare
    \\

    \bottomrule

    \end{tabularx}
    \caption{Italian auxiliary rules. English must have specified lemma. Italian alignment cannot have a lemma in the specified list.}
    \label{tab:italian-aux-rules}
\end{table}

\begin{table*}[h]
    \scriptsize
    \centering
    \begin{tabularx}{\textwidth}{l rrr rrr c rrr}
    \toprule
    &
    \multicolumn{3}{c}{English ($T_e$)} &
    \multicolumn{3}{c}{Spanish ($T_t$)} &
    \multicolumn{1}{c}{Coref English ($C_e$)} &
    \multicolumn{3}{c}{Coref Spanish ($C_t$) }
    \\
    \cmidrule(r){2-4} 
    \cmidrule(r){5-7}
    \cmidrule(r){8-8}
    \cmidrule(r){9-11}
    
    Rule &

    \multicolumn{1}{c}{Form} &
    \multicolumn{1}{c}{POS} &
    \multicolumn{1}{c}{Case} &

    \multicolumn{1}{c}{Form} &
    \multicolumn{1}{c}{POS} &
    \multicolumn{1}{c}{Case} &

    \multicolumn{1}{c}{POS} &

    \multicolumn{1}{c}{POS} &
    \multicolumn{1}{c}{Gender} &
    \multicolumn{1}{c}{Number} \\
    
    \midrule

\texttt{NOM.FEM.SING} & it & \texttt{PNOUN} & Nom. & ella & \texttt{PNOUN} & * & \texttt{NOUN} & \texttt{NOUN} & Fem. & Sing. \\
\texttt{NOM.MASC.SING} & it & \texttt{PNOUN} & Nom. & él & \texttt{PNOUN} & * & \texttt{NOUN} & \texttt{NOUN} & Masc. & Sing. \\
\texttt{NOM.FEM.PLUR} & it & \texttt{PNOUN} & Nom. & ellas & \texttt{PNOUN} & * & \texttt{NOUN} & \texttt{NOUN} & Fem. & Plur. \\
\texttt{NOM.MASC.PLUR} & it & \texttt{PNOUN} & Nom. & ellos & \texttt{PNOUN} & * & \texttt{NOUN} & \texttt{NOUN} & Masc. & Plur. \\
\texttt{ACC.MASC.SING} & it & \texttt{PNOUN} & Acc. & lo & \texttt{PNOUN} & * & \texttt{NOUN} & \texttt{NOUN} & Masc. & Sing. \\
\texttt{ACC.FEM.SING} & it & \texttt{PNOUN} & Acc. & la & \texttt{PNOUN} & * & \texttt{NOUN} & \texttt{NOUN} & Fem. & Sing. \\
\texttt{ACC.MASC.PLUR} & them & \texttt{PNOUN} & Acc. & los & \texttt{PNOUN} & * & \texttt{NOUN} & \texttt{NOUN} & Masc. & Sing. \\
\texttt{ACC.FEM.PLUR} & them & \texttt{PNOUN} & Acc. & las & \texttt{PNOUN} & * & \texttt{NOUN} & \texttt{NOUN} & Fem. & Sing. \\
\texttt{DISJ.MASC.SING} & it & \texttt{PNOUN} & -Nom. & él & \texttt{PNOUN} & * & \texttt{NOUN} & \texttt{NOUN} & Masc. & Sing. \\
\texttt{DISJ.MASC.SING.ALT} & it & \texttt{PNOUN} & -Nom & ello & \texttt{PNOUN} & * & \texttt{NOUN} & \texttt{NOUN} & Masc. & Sing. \\
\texttt{DISJ.FEM.SING} & it & \texttt{PNOUN} & -Nom & ella & \texttt{PNOUN} & * & \texttt{NOUN} & \texttt{NOUN} & Fem. & Sing. \\
\texttt{DISJ.MASC.PLUR} & them & \texttt{PNOUN} & -Nom & ellos & \texttt{PNOUN} & * & \texttt{NOUN} & \texttt{NOUN} & Masc. & Plur. \\
\texttt{DISJ.FEM.PLUR} & them & \texttt{PNOUN} & -Nom & ellas & \texttt{PNOUN} & * & \texttt{NOUN} & \texttt{NOUN} & Fem. & Plur. \\
\texttt{NOM.INFORM.SING} & you & \texttt{PNOUN} & Nom. & tú & \texttt{PNOUN} & * & - & - & - & - \\
\texttt{NOM.FORM.SING} & you & \texttt{PNOUN} & Nom. & usted & \texttt{PNOUN} & * & - & - & - & - \\
\texttt{NOM.FORM.PLUR} & you & \texttt{PNOUN} & Nom. & ustedes & \texttt{PNOUN} & * & - & - & - & - \\
\texttt{NOM.INFORM.PLUR.MASC} & you & \texttt{PNOUN} & Nom. & vosotros & \texttt{PNOUN} & * & - & - & - & - \\
\texttt{NOM.INFORM.PLUR.FEM} & you & \texttt{PNOUN} & Nom. & vosotras & \texttt{PNOUN} & * & - & - & - & - \\
\texttt{ACC.INFORM.SING} & you & \texttt{PNOUN} & Acc. & te & \texttt{PNOUN} & * & - & - & - & - \\
\texttt{ACC.FORM.SING.MASC} & you & \texttt{PNOUN} & Acc. & lo & \texttt{PNOUN} & * & - & - & - & - \\
\texttt{ACC.FORM.SING.FEM} & you & \texttt{PNOUN} & Acc. & la & \texttt{PNOUN} & * & - & - & - & - \\
\texttt{ACC.FORM.PLUR.MASC} & you & \texttt{PNOUN} & Acc. & los & \texttt{PNOUN} & * & - & - & - & - \\
\texttt{ACC.FORM.PLUR.FEM} & you & \texttt{PNOUN} & Acc. & las & \texttt{PNOUN} & * & - & - & - & - \\
\texttt{ACC.INFORM.PLUR} & you & \texttt{PNOUN} & Acc. & os & \texttt{PNOUN} & * & - & - & - & - \\
\texttt{DISJ.INFORM.SING} & you & \texttt{PNOUN} & -Nom. & ti & \texttt{PNOUN} & * & - & - & - & - \\
\texttt{DISJ.INFORM.SING.ALT} & you & \texttt{PNOUN} & -Nom. & contigo & \texttt{PNOUN} & * & - & - & - & - \\
\texttt{DISJ.FORM.SING} & you & \texttt{PNOUN} & -Nom. & usted & \texttt{PNOUN} & * & - & - & - & - \\
\texttt{DISJ.INFORM.PLUR.MASC} & you & \texttt{PNOUN} & -Nom. & vosotros & \texttt{PNOUN} & * & - & - & - & - \\
\texttt{DISJ.INFORM.PLUR.FEM} & you & \texttt{PNOUN} & -Nom. & vosotras & \texttt{PNOUN} & * & - & - & - & - \\
\texttt{DISJ.FORM.PLUR} & you & \texttt{PNOUN} & -Nom. & ustedes & \texttt{PNOUN} & * & - & - & - & - \\

    \bottomrule

    \end{tabularx}
    \caption{Spanish Pronoun Rules}
    \label{tab:spanish-pronoun-rules}
\end{table*}

\begin{table*}[h]
    \scriptsize
    \centering
    \begin{tabularx}{\textwidth}{l rrr rrr c rrr}
    \toprule
    &
    \multicolumn{3}{c}{English ($T_e$)} &
    \multicolumn{3}{c}{French ($T_t$)} &
    \multicolumn{1}{c}{Coref English ($C_e$)} &
    \multicolumn{3}{c}{Coref French ($C_t$) }
    \\
    \cmidrule(r){2-4} 
    \cmidrule(r){5-7}
    \cmidrule(r){8-8}
    \cmidrule(r){9-11}
    
    Rule &

    \multicolumn{1}{c}{Form} &
    \multicolumn{1}{c}{POS} &
    \multicolumn{1}{c}{Case} &

    \multicolumn{1}{c}{Form} &
    \multicolumn{1}{c}{POS} &
    \multicolumn{1}{c}{Case} &

    \multicolumn{1}{c}{POS} &

    \multicolumn{1}{c}{POS} &
    \multicolumn{1}{c}{Gender} &
    \multicolumn{1}{c}{Number} \\
    
    \midrule
\texttt{NOM.FEM.SING} & it & \texttt{PNOUN} & Nom. & elle & \texttt{PNOUN} & * & \texttt{NOUN} & \texttt{NOUN} & Fem. & Sing. \\
\texttt{NOM.MASC.SING} & it & \texttt{PNOUN} & Nom. & il & \texttt{PNOUN} & * & \texttt{NOUN} & \texttt{NOUN} & Masc. & Sing. \\
\texttt{NOM.FEM.PLUR} & they & \texttt{PNOUN} & Nom. & elles & \texttt{PNOUN} & * & \texttt{NOUN} & \texttt{NOUN} & Fem. & Plur. \\
\texttt{NOM.MASC.PLUR} & they & \texttt{PNOUN} & Nom. & ils & \texttt{PNOUN} & * & \texttt{NOUN} & \texttt{NOUN} & Masc. & Plur. \\
\texttt{ACC.MASC.SING} & it & \texttt{PNOUN} & Acc. & le & \texttt{PNOUN} & * & \texttt{NOUN} & \texttt{NOUN} & Masc. & Sing. \\
\texttt{ACC.FEM.SING} & it & \texttt{PNOUN} & Acc. & la & \texttt{PNOUN} & * & \texttt{NOUN} & \texttt{NOUN} & Fem. & Sing. \\
\texttt{GEN.FEM.SING.1S} & mine & \texttt{PNOUN} & * & mienne & \texttt{PNOUN} & * & \texttt{NOUN} & \texttt{NOUN} & Fem. & Sing. \\
\texttt{GEN.FEM.SING.1P} & ours & \texttt{PNOUN} & * & la nôtre & \texttt{PNOUN} & * & \texttt{NOUN} & \texttt{NOUN} & Fem. & Sing. \\
\texttt{GEN.FEM.SING.2S} & yours & \texttt{PNOUN} & * & tienne & \texttt{PNOUN} & * & \texttt{NOUN} & \texttt{NOUN} & Fem. & Sing. \\
\texttt{GEN.FEM.SING.2P} & yours & \texttt{PNOUN} & * & la vôtre & \texttt{PNOUN} & * & \texttt{NOUN} & \texttt{NOUN} & Fem. & Sing. \\
\texttt{GEN.FEM.SING.3SM} & his & \texttt{PNOUN} & * & sienne & \texttt{PNOUN} & * & \texttt{NOUN} & \texttt{NOUN} & Fem. & Sing. \\
\texttt{GEN.FEM.SING.3SF} & hers & \texttt{PNOUN} & * & sienne & \texttt{PNOUN} & * & \texttt{NOUN} & \texttt{NOUN} & Fem. & Sing. \\
\texttt{GEN.FEM.SING.3N} & its & \texttt{PNOUN} & * & sienne & \texttt{PNOUN} & * & \texttt{NOUN} & \texttt{NOUN} & Fem. & Sing. \\
\texttt{GEN.FEM.SING.3P} & theirs & \texttt{PNOUN} & * & la leur & \texttt{PNOUN} & * & \texttt{NOUN} & \texttt{NOUN} & Fem. & Sing. \\
\texttt{NOM.INFORM.SING} & you & \texttt{PNOUN} & Nom. & tu & \texttt{PNOUN} & * & - & - & - & - \\
\texttt{NOM.FORM+PLUR} & you & \texttt{PNOUN} & Nom. & vous & \texttt{PNOUN} & * & - & - & - & - \\
\texttt{ACC.INFORM.SING} & you & \texttt{PNOUN} & Acc. & te & \texttt{PNOUN} & * & - & - & - & - \\
\texttt{ACC.INFORM.SING.LIAS} & you & \texttt{PNOUN} & Acc. & t' & \texttt{PNOUN} & * & - & - & - & - \\
\texttt{ACC.FORM+PLUR} & you & \texttt{PNOUN} & Acc. & vous & \texttt{PNOUN} & * & - & - & - & - \\
\texttt{DISJ.INFORM.SING} & you & \texttt{PNOUN} & -Nom & toi & \texttt{PNOUN} & * & - & - & - & - \\

    \bottomrule

    \end{tabularx}
    \caption{A sampling of French pronoun rules (abridged). Some forms left off for space.}
    \label{tab:french-pronoun-rules}
\end{table*}

\begin{table*}[h]
    \scriptsize
    \centering
    \begin{tabularx}{\textwidth}{l rrr rrr c rrr}
    \toprule
    &
    \multicolumn{3}{c}{English ($T_e$)} &
    \multicolumn{3}{c}{Italian ($T_t$)} &
    \multicolumn{1}{c}{Coref English ($C_e$)} &
    \multicolumn{3}{c}{Coref Italian ($C_t$) }
    \\
    \cmidrule(r){2-4} 
    \cmidrule(r){5-7}
    \cmidrule(r){8-8}
    \cmidrule(r){9-11}
    
    Rule &

    \multicolumn{1}{c}{Form} &
    \multicolumn{1}{c}{POS} &
    \multicolumn{1}{c}{Case} &

    \multicolumn{1}{c}{Form} &
    \multicolumn{1}{c}{POS} &
    \multicolumn{1}{c}{Case} &

    \multicolumn{1}{c}{POS} &

    \multicolumn{1}{c}{POS} &
    \multicolumn{1}{c}{Gender} &
    \multicolumn{1}{c}{Number} \\
    
    \midrule

\texttt{NOM.MASC.SING} & it & \texttt{PNOUN} & Nom. & lui & \texttt{PNOUN} & * & \texttt{NOUN} & \texttt{NOUN} & Masc. & Sing. \\
\texttt{NOM.FEM.SING} & it & \texttt{PNOUN} & Nom. & lei & \texttt{PNOUN} & * & \texttt{NOUN} & \texttt{NOUN} & Fem. & Sing. \\
\texttt{ACC.MASC.SING} & it & \texttt{PNOUN} & Acc. & lo & \texttt{PNOUN} & * & \texttt{NOUN} & \texttt{NOUN} & Masc. & Sing. \\
\texttt{ACC.FEM.SING} & it & \texttt{PNOUN} & Acc. & la & \texttt{PNOUN} & * & \texttt{NOUN} & \texttt{NOUN} & Fem. & Sing. \\
\texttt{ACC.MASC.PLUR} & them & \texttt{PNOUN} & Acc. & li & \texttt{PNOUN} & * & \texttt{NOUN} & \texttt{NOUN} & Masc. & Plur. \\
\texttt{ACC.FEM.PLUR} & them & \texttt{PNOUN} & Acc. & le & \texttt{PNOUN} & * & \texttt{NOUN} & \texttt{NOUN} & Fem. & Plur. \\
\texttt{DAT.MASC.SING} & it & \texttt{PNOUN} & Acc. & gli & \texttt{PNOUN} & * & \texttt{NOUN} & \texttt{NOUN} & Masc. & Sing. \\
\texttt{DAT.FEM.SING} & it & \texttt{PNOUN} & Acc. & le & \texttt{PNOUN} & * & \texttt{NOUN} & \texttt{NOUN} & Fem. & Sing. \\
\texttt{DISJ.MASC.SING} & it & \texttt{PNOUN} & -Nom & lui & \texttt{PNOUN} & * & \texttt{NOUN} & \texttt{NOUN} & Masc. & Sing. \\
\texttt{DISJ.FEM.SING} & it & \texttt{PNOUN} & -Nom & lei & \texttt{PNOUN} & * & \texttt{NOUN} & \texttt{NOUN} & Fem. & Sing. \\
\texttt{GEN.FEM.SING.1S} & mine & \texttt{PNOUN} & * & mia & \texttt{PNOUN} & * & \texttt{NOUN} & \texttt{NOUN} & Fem. & Sing. \\
\texttt{GEN.FEM.SING.2S} & yours & \texttt{PNOUN} & * & tua & \texttt{PNOUN} & * & \texttt{NOUN} & \texttt{NOUN} & Fem. & Sing. \\
\texttt{GEN.FEM.SING.3M} & his & \texttt{PNOUN} & * & sua & \texttt{PNOUN} & * & \texttt{NOUN} & \texttt{NOUN} & Fem. & Sing. \\
\texttt{GEN.FEM.SING.3F} & hers & \texttt{PNOUN} & * & sua & \texttt{PNOUN} & * & \texttt{NOUN} & \texttt{NOUN} & Fem. & Sing. \\
\texttt{GEN.FEM.SING.3N} & its & \texttt{PNOUN} & * & sua & \texttt{PNOUN} & * & \texttt{NOUN} & \texttt{NOUN} & Fem. & Sing. \\
\texttt{GEN.FEM.SING.2P} & yours & \texttt{PNOUN} & * & vostra & \texttt{PNOUN} & * & \texttt{NOUN} & \texttt{NOUN} & Fem. & Sing. \\
\texttt{GEN.FEM.SING.3P} & theirs & \texttt{PNOUN} & * & loro & \texttt{PNOUN} & * & \texttt{NOUN} & \texttt{NOUN} & Fem. & Sing. \\
\texttt{NOM.INFORM.SING} & you & \texttt{PNOUN} & * & tu & \texttt{PNOUN} & * & - & - & - & - \\
\texttt{NOM.FORM.SING} & you & \texttt{PNOUN} & * & lei & \texttt{PNOUN} & * & - & - & - & - \\
\text{NOM.INFORM.PLUR} & you & \texttt{PNOUN} & * & voi & \texttt{PNOUN} & * & - & - & - & - \\

    \bottomrule
     
    \end{tabularx}
    \caption{A sampling of Italian Pronoun Rules. We do not consider the conflated Italian pronouns which combine accusatives and datives which co-occur. English case is used as it is a better model. Accusative is used for dative since the SpaCy models conflate the two in English.}
    \label{tab:italian-pronoun-rules}
\end{table*}

\begin{table*}[h]
    \scriptsize
    \centering
    \begin{tabularx}{\textwidth}{l rrr rrr c rrr}
    \toprule
    &
    \multicolumn{3}{c}{English ($T_e$)} &
    \multicolumn{3}{c}{Polish ($T_t$)} &
    \multicolumn{1}{c}{Coref English ($C_e$)} &
    \multicolumn{3}{c}{Coref Polish ($C_t$) }
    \\
    \cmidrule(r){2-4} 
    \cmidrule(r){5-7}
    \cmidrule(r){8-8}
    \cmidrule(r){9-11}
    
    Rule &

    \multicolumn{1}{c}{Form} &
    \multicolumn{1}{c}{POS} &
    \multicolumn{1}{c}{Case} &

    \multicolumn{1}{c}{Form} &
    \multicolumn{1}{c}{POS} &
    \multicolumn{1}{c}{Case} &

    \multicolumn{1}{c}{POS} &

    \multicolumn{1}{c}{POS} &
    \multicolumn{1}{c}{Gender} &
    \multicolumn{1}{c}{Number} \\
    
    \midrule
\texttt{NOM.NEUT.SING} & it & \texttt{PNOUN} & * & ono & \texttt{PNOUN} & Nom. & \texttt{NOUN} & \texttt{NOUN} & Neut. & Sing. \\
\texttt{NOM.MASC.SING} & it & \texttt{PNOUN} & * & on & \texttt{PNOUN} & Nom. & \texttt{NOUN} & \texttt{NOUN} & Masc. & Sing. \\
\texttt{NOM.FEM.SING} & it & \texttt{PNOUN} & * & ona & \texttt{PNOUN} & Nom. & \texttt{NOUN} & \texttt{NOUN} & Fem. & Sing. \\
\texttt{ACC.NEUT.SING} & it & \texttt{PNOUN} & * & je & \texttt{PNOUN} & Acc. & \texttt{NOUN} & \texttt{NOUN} & Neut. & Sing. \\
\texttt{ACC.NEUT.SING.ALT1} & it & \texttt{PNOUN} & * & nie & \texttt{PNOUN} & Acc. & \texttt{NOUN} & \texttt{NOUN} & Neut. & Sing. \\
\texttt{ACC.MASC.SING} & it & \texttt{PNOUN} & * & je & \texttt{PNOUN} & Acc. & \texttt{NOUN} & \texttt{NOUN} & Masc. & Sing. \\
\texttt{ACC.MASC.SING.ALT} & it & \texttt{PNOUN} & * & niego & \texttt{PNOUN} & Acc. & \texttt{NOUN} & \texttt{NOUN} & Masc. & Sing. \\
\texttt{ACC.FEM.SING} & it & \texttt{PNOUN} & * & ją & \texttt{PNOUN} & Acc. & \texttt{NOUN} & \texttt{NOUN} & Fem. & Sing. \\
\texttt{GEN.NEUT.SING} & it & \texttt{PNOUN} & * & jego & \texttt{PNOUN} & Gen. & \texttt{NOUN} & \texttt{NOUN} & Neut. & Sing. \\
\texttt{GEN.NEUT.SING.ALT1} & it & \texttt{PNOUN} & * & niego & \texttt{PNOUN} & Gen. & \texttt{NOUN} & \texttt{NOUN} & Neut. & Sing. \\
\texttt{GEN.NEUT.SING.ALT2} & it & \texttt{PNOUN} & * & go & \texttt{PNOUN} & Gen. & \texttt{NOUN} & \texttt{NOUN} & Neut. & Sing. \\
\texttt{GEN.MASC.SING} & it & \texttt{PNOUN} & * & je & \texttt{PNOUN} & Gen. & \texttt{NOUN} & \texttt{NOUN} & Masc. & Sing. \\
\texttt{GEN.MASC.SING.ALT1} & it & \texttt{PNOUN} & * & niego & \texttt{PNOUN} & Gen. & \texttt{NOUN} & \texttt{NOUN} & Masc. & Sing. \\
\texttt{GEN.FEM.SING} & it & \texttt{PNOUN} & * & jej & \texttt{PNOUN} & Gen. & \texttt{NOUN} & \texttt{NOUN} & Fem. & Sing. \\
\texttt{GEN.FEM.SING.ALT1} & it & \texttt{PNOUN} & * & niej & \texttt{PNOUN} & Gen. & \texttt{NOUN} & \texttt{NOUN} & Fem. & Sing. \\
\texttt{LOC.NEUT.SING} & it & \texttt{PNOUN} & * & nim & \texttt{PNOUN} & Loc. & \texttt{NOUN} & \texttt{NOUN} & Neut. & Sing. \\
\texttt{LOC.MASC.SING} & it & \texttt{PNOUN} & * & nim & \texttt{PNOUN} & Loc. & \texttt{NOUN} & \texttt{NOUN} & Masc. & Sing. \\
\texttt{LOC.FEM.SING} & it & \texttt{PNOUN} & * & niej & \texttt{PNOUN} & Loc. & \texttt{NOUN} & \texttt{NOUN} & Fem. & Sing. \\
\texttt{DAT.NEUT.SING} & it & \texttt{PNOUN} & * & jemu & \texttt{PNOUN} & Dat. & \texttt{NOUN} & \texttt{NOUN} & Neut. & Sing. \\
\texttt{INS.NEUT.SING} & it & \texttt{PNOUN} & * & nim & \texttt{PNOUN} & Ins. & \texttt{NOUN} & \texttt{NOUN} & Neut. & Sing. \\
\texttt{NOM.INFORM.SING} & you & \texttt{PNOUN} & * & ty & \texttt{PNOUN} & Nom. & - & - & - & - \\
\texttt{ACC.INFORM.SING} & you & \texttt{PNOUN} & * & ciebie & \texttt{PNOUN} & Acc. & - & - & - & - \\
\texttt{NOM.FORM.SING.FEM} & you & \texttt{PNOUN} & * & pani & \texttt{PNOUN} & Nom. & - & - & - & - \\
\texttt{ACC.FORM.SING.FEM} & you & \texttt{PNOUN} & * & panią & \texttt{PNOUN} & Acc. & - & - & - & - \\

    \bottomrule

    \end{tabularx}
    \caption{A sampling of Polish Pronoun Rules. Some left off for space.}
    \label{tab:polish-pronoun-rules}
\end{table*}

\begin{table*}[ht]
        \small
        \centering
\setlength{\tabcolsep}{3pt}
    \begin{tabular}{ll rr rr rr r}

    \toprule
    & 
    & \multicolumn{1}{c}{DE}
    & \multicolumn{1}{c}{ES}
    & \multicolumn{1}{c}{FR}
    & \multicolumn{1}{c}{IT}
    & \multicolumn{1}{c}{PL}
    & \multicolumn{1}{c}{PT}
    & \multicolumn{1}{c}{RU} \\
    \midrule

    \multirow{3}{*}{\gender} & sent. & 29.0 & 35.4 & 32.6 & 28.7 & 23.8 & 27.8 & 24.7 \\
    & doc. & \textbf{33.8} & \textbf{38.7} & \textbf{37.2} & \textbf{32.7} & \textbf{27.1} & \textbf{31.3} & \textbf{27.6} \\
    & &  +4.8 & +4.6 & +2.9 & +3.3 & +3.5 & +4.0 & +3.3\\

    \midrule

    \multirow{3}{*}{\animacy} & sent. 
    & 33.3 & 44.3 & 37.5 & 35.1 & 29.8 & 40.5 & 32.1
    \\
    & doc. 
    & \textbf{37.7} & \textbf{48.3} & \textbf{40.6} & \textbf{37.6} & \textbf{32.3} & \textbf{44.4} & \textbf{36.0}
    \\
    & 
    & +4.4 & +4.0 & +3.1 & +2.5 & +2.5 & +3.9 & +3.9
    \\

    \midrule

    \multirow{3}{*}{\formality} & sent.
    & 26.4 & 32.0 & 28.4 & 21.7 & 36.1 & 29.2 & 34.3
    \\
    & doc.
    & \textbf{28.4} & \textbf{35.6} & \textbf{30.2} & \textbf{23.4} & \textbf{37.1} & \textbf{31.3} & \textbf{36.1}
    \\
    & 
    & +2.0 & +3.6 & +1.8 & +1.7 & +1.0 & +2.1 & +1.8
    \\

    \midrule
    \multirow{3}{*}{\auxiliary} & sent.
    & 17.7 & 17.3 & 14.9 & 17.8 & 15.3 & 15.8 & 19.9  \\
    & doc.
    & \textbf{30.1} & \textbf{33.4} & \textbf{33.6} & \textbf{34.7} & \textbf{33.1} & \textbf{32.5} & \textbf{42.2}
    \\
    & 
    & +12.4 & +16.1 & +18.7 & +16.9 & +17.8 & +16.7 & +22.3
    \\

    \midrule

    \multirow{3}{*}{\inflection} & sent.
    & - & - & - & - & 27.3 & - & 27.7 \\
    & doc. & - & - & - & - & \textbf{30.7} & - & \textbf{29.9} \\
    & & - & - & - & - & +2.2 & - & +3.4 \\
    \bottomrule

    \end{tabular}
\caption{BLEU scores to evaluate the translation quality of this model. Higher is better. \emph{sent.} denotes that no additional context was given while \emph{doc.} was given five consecutive sentences. All translations produced by DeepL commercial API.}
\label{tab:bleu-alltables}

\end{table*}

\end{document}